\pdfoutput=1

\documentclass[11pt]{article}

\usepackage[table]{xcolor} 

\usepackage[preprint]{acl}

\usepackage{times}
\usepackage{latexsym}

\usepackage[T1]{fontenc}

\usepackage[utf8]{inputenc}

\usepackage{microtype}

\usepackage{inconsolata}

\usepackage{graphicx}

\usepackage[utf8]{inputenc} 
\usepackage[T1]{fontenc}    
\usepackage{hyperref}       
\usepackage{url}            
\usepackage{booktabs}       
\usepackage{amsfonts}       
\usepackage{nicefrac}       
\usepackage{microtype}      
\usepackage{xcolor}         
\usepackage{graphicx}
\usepackage{algorithmic}
\usepackage{multirow}
\usepackage[linesnumbered,ruled,vlined]{algorithm2e}
\usepackage{amssymb}
\usepackage{cite}
\usepackage{amsmath}

\usepackage{color}
\usepackage{colortbl}      
\usepackage[most]{tcolorbox}

\usepackage{subcaption}
\usepackage{makecell}
\usepackage[symbol]{footmisc}
\usepackage{fancyhdr}
\usepackage{hyperref}

\definecolor{myred}{HTML}{C00000}
\definecolor{myblue}{HTML}{002060}

\newcommand{\ourmethod}{D2R }

\title{Bridging the Dynamic Perception Gap: Training-Free Draft Chain-of-Thought for Dynamic Multimodal Spatial Reasoning}

\author{Siqu Ou \textsuperscript{\rm 1,}\thanks{Equal contribution} \quad
  Hongcheng Liu \textsuperscript{\rm 1,}\footnotemark[1] \quad
  Pingjie Wang  \textsuperscript{\rm 1,2} \quad
  Yusheng Liao  \textsuperscript{\rm 1,2}  \\
  {\bf Chuan Xuan  \textsuperscript{\rm 1}} \quad
  {\bf Yanfeng Wang \textsuperscript{\rm 1,2}} \quad
  {\bf Yu Wang \textsuperscript{\rm 1,2,}}\thanks{Corresponding author}\\
    \textsuperscript{\rm 1} Shanghai Jiao Tong University \quad
    \textsuperscript{\rm 2} Shanghai AI lab
  }

\begin{document}
\maketitle

\begin{abstract}

While chains-of-thought (CoT) have advanced complex reasoning in multimodal large language models (MLLMs), existing methods remain confined to text or static visual domains, often faltering in dynamic spatial reasoning tasks. To bridge this gap, we present GRASSLAND, a novel maze navigation benchmark designed to evaluate dynamic spatial reasoning. Our experiments show that augmenting textual reasoning chains with dynamic visual drafts, overlaid on input images, significantly outperforms conventional approaches, offering new insights into spatial reasoning in evolving environments. To generalize this capability, we propose D2R (Dynamic Draft-Augmented Reasoning), a training-free framework that seamlessly integrates textual CoT with corresponding visual drafts into MLLMs. Extensive evaluations demonstrate that D2R consistently enhances performance across diverse tasks, establishing a robust baseline for dynamic spatial reasoning without requiring model fine-tuning. Project is open at \url{https://github.com/Cratileo/D2R}.

\end{abstract}

\section{Introduction}

Humans often exhibit effective behavioral strategies that inspire multimodal large language models (MLLMs)~\citep{gpt4v, llava, deepseekvl2, minicpmv} to tackle complex tasks, particularly in the realm of multimodal reasoning. In such tasks, humans commonly create drafts to support step-by-step thinking when processing visual information that integrates text and imagery. This drafting approach is especially beneficial for extracting insights from dynamic images, where chronological, incremental reasoning is highly effective.

\begin{figure}[t]
    \centering    
    \includegraphics[width=\linewidth]{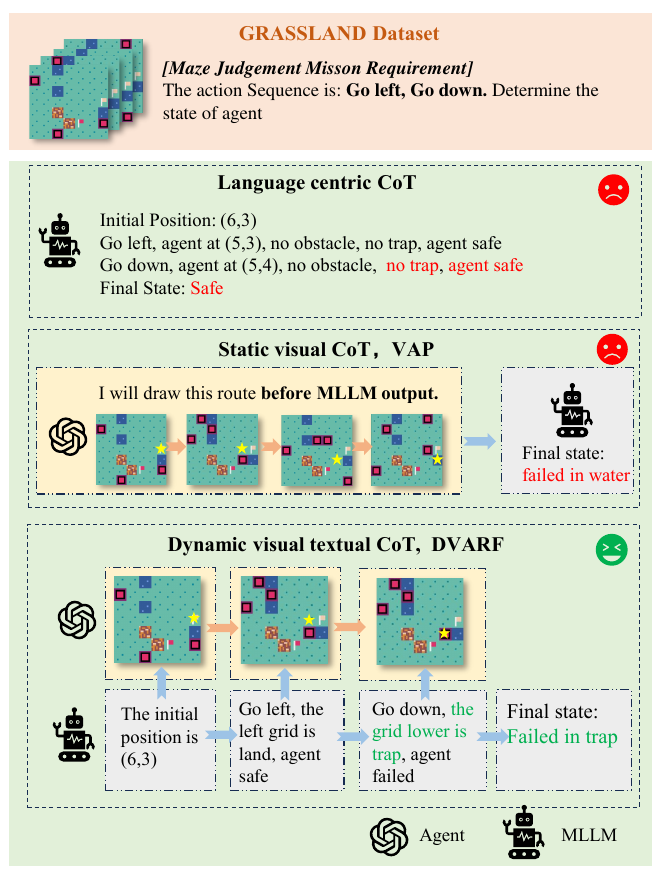}
    \caption{The demonstration of the Draft CoT with D2R. Compared to the spatial information gaps in language-centric CoT, and the incomplete dynamic information in static visual CoT, which only visualizes the input rather than the MLLM’s thought process, Draft CoT excels at dynamic spatial reasoning. }
    \label{fig:case-1}
\end{figure}

Current MLLMs primarily emphasize step-by-step reasoning patterns or simple visualization techniques, exemplified by methods such as ToT~\citep{tot} and ICoT~\citep{icot}, but they lack mechanisms for draft creation based on input images. While these frameworks achieve strong results on textual and static visual tasks~\citep{m3cot, math, mmecot, data1}, they often suffer from loss of rich visual information and diminished spatial awareness—factors critical for dynamic multimodal spatial reasoning. Since dynamic spatial reasoning plays a pivotal role in many real-world applications, it is important to investigate how well existing models perform in this domain.

To address this, we develop GRASSLAND, a dynamic maze environment modeled as a classical pixel grid world with evolving environment grids. We define two dynamic spatial reasoning tasks—Maze Judgment and Maze Navigation—to evaluate models’ ability to perform complex visual analysis in changing contexts. As illustrated in Figure~\ref{fig:case-1}, our experiments reveal that existing MLLMs and reasoning frameworks struggle with these tasks, often overlooking or misinterpreting spatial context, such as inaccurately judging locations or ignoring special grid features. To overcome these challenges, we propose the Draft Chain-of-Thought (Draft CoT) approach, which integrates textual reasoning with corresponding drafts over dynamic input images. This method significantly outperforms previous approaches, providing fresh insights into dynamic spatial reasoning.

\begin{figure*}[t]
    \centering
    \includegraphics[width=.9\textwidth, trim=0cm 0cm 0cm 0cm, clip]{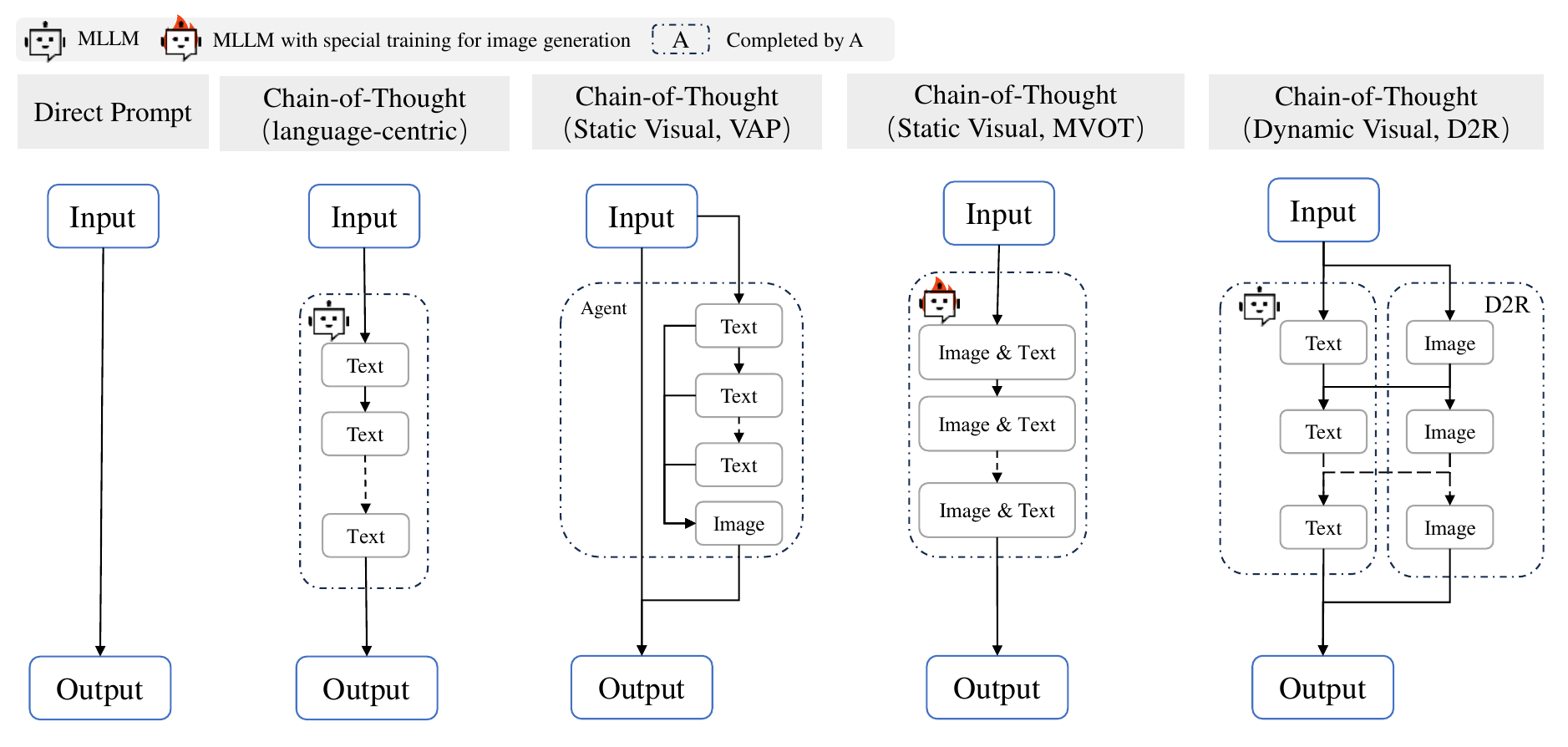}
    \caption{Illustration of the difference between our method and others. Direct prompting and language-centric CoT face significant limitations in dynamic spatial reasoning tasks without images. VAP can only generate static images based on agent prompts, without MLLM involvement for dynamic perception. MVOT requires MLLMs powerful in image generation by training on specialized datasets. In contrast, D2R marks the textual thought in the image as draft and integrates it into the Draft CoT, enhancing the MLLM's dynamic spatial reasoning ability without specific training.}
    \label{fig:main}
\end{figure*}

Despite its effectiveness, Draft CoT relies on image generation capabilities not universally available across all MLLMs. To broaden its applicability, we introduce a training-free framework named the \textbf{Dynamic Draft Augmented Reasoning Framework} (D2R). As shown in Figure~\ref{fig:main}, D2R seamlessly integrates both visual and textual inputs, enhancing reasoning by enabling cross-modal information exchange. It first generates a global plan based on the task prompt and tool set, then iteratively performs chronological reasoning by updating textual thoughts as drafts on dynamic images. Finally, D2R signals the MLLM to produce the final output, concluding the iterative process.

Extensive experiments on the two dynamic spatial reasoning tasks demonstrate that D2R surpasses existing text-only and static vision-based reasoning methods. Moreover, tests on multiple MLLMs confirm D2R’s ease of transfer, robustness, and broad applicability as a training-free enhancement.

\begin{figure*}[htbp]
    \centering
    \includegraphics[width=.98\textwidth]{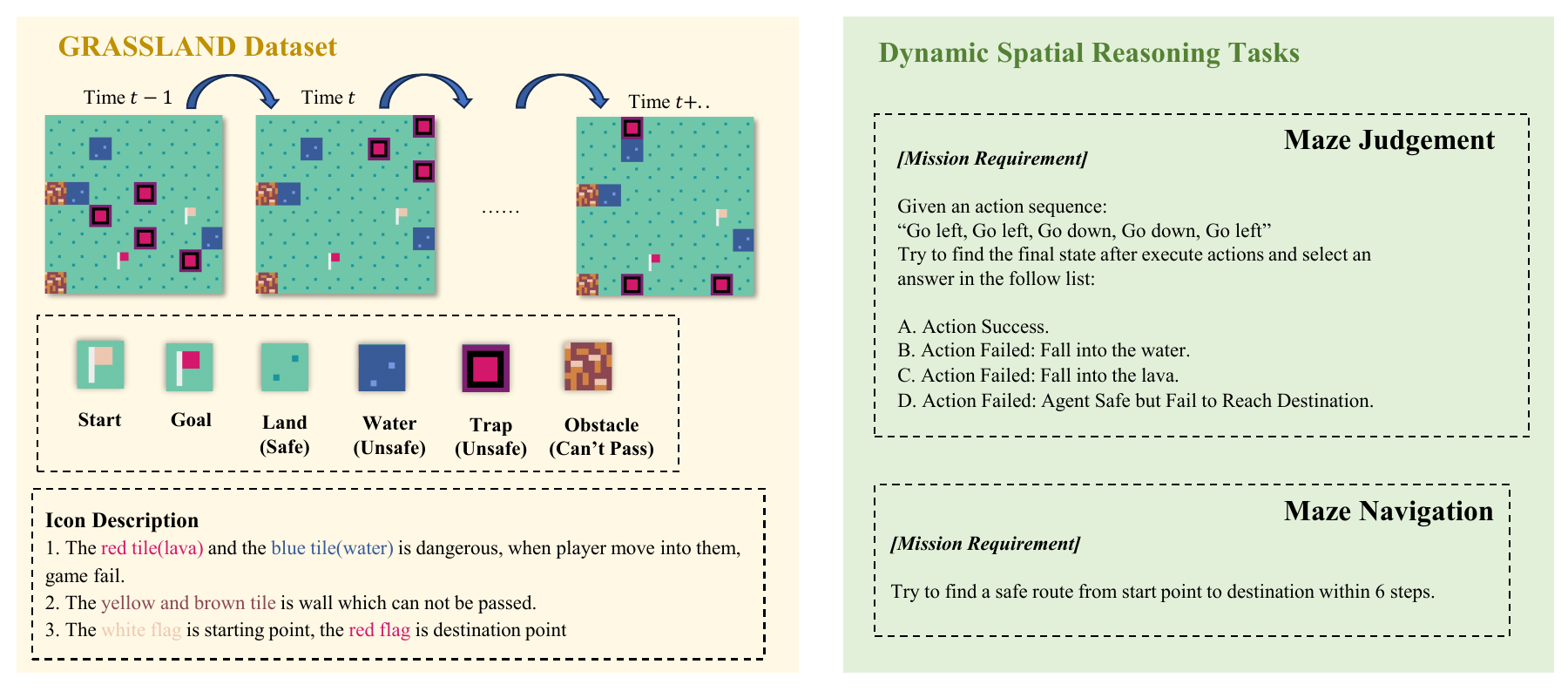}
    \caption{Example of dynamic scenario sequence in GRASSLAND. The left part is the illustration of the dynamic images and grids in GRASSLAND, and the right part is the description of the two tasks.}
    \label{fig:map}
\end{figure*}

In summary, this paper makes three main contributions:
\begin{itemize}
\item \textbf{A novel benchmark for dynamic spatial reasoning:} We introduce GRASSLAND, a classical pixel grid world with dynamic environment changes, along with two challenging tasks—Maze Judgment and Maze Navigation—to rigorously evaluate dynamic spatial reasoning capabilities.
\item \textbf{A new Draft Chain-of-Thought method:} We propose Draft CoT, which combines textual reasoning with corresponding drafts over dynamic input images, significantly improving performance over existing reasoning frameworks on dynamic spatial tasks.
\item \textbf{A training-free framework for broad applicability:} We develop the Dynamic Draft Augmented Reasoning Framework (D2R) that seamlessly integrates Draft CoT into existing MLLMs without additional training, enabling enhanced dynamic multimodal reasoning across various models.
\end{itemize}

\section{Related Works}

\subsection{MultiModal Large Language Models}
Multimodal Large Language Models~\citep{nvila, qwen2.5o, internvl3} have made remarkable progress by integrating various modalities—such as text, images, and video—into a unified framework for understanding and reasoning. In this framework, different modality encoders project inputs into a shared semantic space, which is then processed by a language model to generate responses~\citep{Yin_2024}. However, most existing MLLMs adopt a unimodal generation strategy: they rely solely on text for auto-regressive response generation, treating non-text modalities merely as auxiliary context during encoding~\citep{Liu2024MedPMCMP}. As a result, the rich and dynamic information contained in modalities like images and videos is not fully utilized during the generation process, which significantly limits the model’s performance on multimodal reasoning tasks~\citep{MSG-BART}. In contrast, OpenAI o3~\citep{o3} demonstrates the potential of step-by-step generation that jointly conditions on both visual and textual inputs. Unfortunately, current MLLMs are not capable of this generation pattern due to inherent limitations in image processing. In this paper, we propose a Dynamic Draft Augmented Reasoning Framework, which achieves adaptiveness-enhanced reasoning with multiple domain inputs by utilizing external tools to generate a bimodal chain-of-thought.

\subsection{MLLMs Reasoning}

Multimodal reasoning tasks are designed to evaluate the ability to integrate information from different modalities and perform comprehensive reasoning~\citep{icot,ddcot}. The most common method is the language-centric multimodal reasoning pattern, which focuses on extracting information from the visual modality and downscaling it to the linguistic domain for inference~\citep{mmreact,llavacot,relate1}. 
Rather from the language-centric pattern, the collaborative multimodal reasoning introduces the visual domain into the reasoning process, such as VAP~\citep{vap} and MVoT~\citep{mvot}. However, VAP merely visualizes the input of the model instead of the model's thought process, while MVoT requires the model to generate multimodal output. Both methods overlooks the need to enhance the generalization ability of existing models across multimodal reasoning tasks. In this paper, we propose Dynamic Draft Augmented Reasoning Framework, which enhances the reasoning capabilities of existing MLLMs by realizing bimodal chains of thought through the combination of textual thought and their corresponding drafts in the input images.

\section{Dynamic Multimodal Spatial Reasoning}

To further evaluate the performance of the existing MLLMs on the dynamic spatial reasoning task, we propose GRASSLAND, a dynamic maze navigation scenario for the dynamic spatial reasoning task.
As shown in Figure~\ref{fig:map}, it simulates a classical pixel grid world $W$ with a start point $p_s$ and destination point $p_e$. Also, parts of the environment grids contain obstacles(‘the walls’)$P_o$, dynamic traps (‘the lava’)$P_l$, and stationary traps (‘the water’) $P_w$. 
The model is required to determine the next action or state based on the given prompt and scenario.

\subsection{Task Formulation}
Based on this dataset, we define two scenarios for the dynamic spatial reasoning tasks: Maze Judgment and Maze Navigation. These scenarios require models to analyze time and spatial sequences, locate special objects, make action decisions, and predict states when actions are executed. The details are presented in Table \ref{tab:dataA}.

\begin{table}[t]
    \centering
    \resizebox{.48\textwidth}{!}{%
    \begin{tabular}{lccclccc}
    \toprule
        \multirow{2}{*}{Task} & \multicolumn{3}{c}{Maze Judgment} & & \multicolumn{3}{c}{Maze planning} \\ 
        \cline{2-4} \cline{6-8} & easy & normal & hard & & easy & normal & hard\\ \midrule
        Grid Size & \multicolumn{3}{c}{$7\times7$} & & \multicolumn{3}{c}{$5\times 5$} \\ 
        Obstacles & 0 & 1 & 2 & & 1 & 2 & 3\\ 
        Dynamic Trap & 2 & 3 & 4 & & 1 & 2& 2\\
        Static Trap & 0 & 1 & 2 & & 0-4 & 0-4 & 0-6\\
        Route Length & 5.32 & 6.00 & 5.67 & & 3.47 & 3.75 & 4.34 \\ 
        \bottomrule
    \end{tabular}}
    \caption{Statistics of the dataset information, covering three levels of complexity in two tasks.}
    \label{tab:dataA}
\end{table}


\paragraph{The Maze Judgment Scenario}  To assess the ability of MLLMs to perceive dynamic spatial locations, we introduce the maze judgment scenario. In this task, the MLLM must determine the final state based on actions and the map, which are divided into success, failure, and loss. This process is modeled within a discrete state space, \( S \), where each state \( s_t \in S \) represents the agent's status at time \( t \). In practice, the model must predict the state in time $t$ defined as $s_{t}$, and determine the final state $s_{end}$, given a world map \( W \) and a sequence of actions \( R_{\text{action}} = \{r_1, r_2, \dots , r_T\} \). This process is performed as follows:

\begin{equation}\label{eq:mjs}
    \fontsize{10}{12}
    \selectfont
    s_t = f(W, R_{action<t}, S_{<t}) \quad t \in \{1,\dots,T\},
\end{equation}
\begin{equation}\label{eq:mjs1}
    s_{end} = s_T.
\end{equation}

\paragraph{The Maze Navigation Scenario}  To examine the ability of MLLM to reason dynamic spatial location, we propose the maze navigation scenario.
In this task, the MLLM should reach the destination from the starting point, while avoiding all dangers and doing so as quickly as possible. This route is defined with the current position $p_t$ and next action $r_t$. In practice, MLLM should lay out a safe route $R_{action}$ that can stay out of danger positions set $P_D= P_l \cup P_w$ (i.e., $\forall t < T,  p_t \notin P_D$), and reach the destination $p_e$ within a limited steps $L$ (i.e., $T \le L$). This process is performed as follows:



\begin{equation}
    \fontsize{10}{12}
    \selectfont
    \label{eq:mns}
     r_t, p_t  = f(W, r_{t-1}, p_{t-1}), \forall t \in \{1,\dots,T\}
\end{equation}
\begin{equation}\label{eq:mns2}
    R_{action} = \{r_t\}_{t=1}^T
\end{equation}
If the agent cannot reach the final destination within a limited steps or fall into the danger set, the agent will be judged as a failure in this case.


\subsection{Interesting Findings}
\begin{table*}[t]
    \centering
    \selectfont
    \begin{tabular}{lccclccc}
    \toprule
        \multirow{2}{*}{Model} & \multicolumn{3}{c}{Maze Judgment} & & \multicolumn{3}{c}{Maze Navigation} \\ 
        \cline{2-4}  \cline{6-8}
        & easy & normal & hard & & easy & normal & hard\\ \midrule
        VideoLLaMA3-7B~\citep{VideoLLaMA3}  & 18.0  & 12.5  & 11.0 & & 1.0  & 1.5  & 0.0  \\ 
        Qwen2.5VL-7B~\citep{Qwen2.5-VL}  & 22.5  & \underline{34.0}  & \underline{28.5} & & 1.0  & 2.0  & 1.0  \\ 
        InternVL2.5-8B~\citep{internVL2.5}  & 21.0  & 18.5  & 19.5 & & 3.5  & 1.0  & 0.5  \\ 
        Qwen2.5VL-32B~\citep{Qwen2.5-VL}  & 14.0  & 9.0  & 9.0 & & 0.0  & 0.0  & 0.0  \\ 
        InternVL2.5-38B~\citep{internVL2.5}   & 22.5  & 26.0  & 25.0 & & 13.5  & \underline{11.5}  & \underline{3.5}  \\
        Qwen2.5VL-72B~\citep{Qwen2.5-VL}  & \textbf{61.0}  & \textbf{38.5}  & 19.0 & & \textbf{31.0}  & \textbf{21.5}  & \textbf{6.5}\\
        InternVL2.5-78B~\citep{internVL2.5}   & 28.5  & 26.0  & \textbf{29.5} & & 15.0  & 9.0  & 1.5  \\ 
        QwenVL-Max~\citep{Qwen-VL} & \underline{40.0} & 21.5 & 14.0 & & \underline{19.5} & 10.5 & 1.5 \\
        \bottomrule
    \end{tabular}
    \caption{Performance of various models in Maze Judgment task and Maze Navigation task with direct prompt. The best results of each dimension are \textbf{bold} and the secondary results are \underline{underlined}.}
    \label{tab:analysis_ab}
\end{table*}

\paragraph{Poor abilities of MLLMs}
To explore the abilities of MLLMs on dynamic spatial reasoning, we measured two tasks on different MLLMs. As shown in Table~\ref{tab:analysis_ab}, MLLM exhibits a poor ability to follow the long action sequence and collaborative processing of information across multiple modalities. Among the failed cases, we note that MLLMs often ignore or misjudge the scenario context in their thinking process, such as misjudging the location or ignoring special grids. These findings suggest that current MLLMs lack a robust mechanism for integrating spatial and contextual cues over time.

\begin{figure}[b]
    \centering
    \includegraphics[width=.95\linewidth]{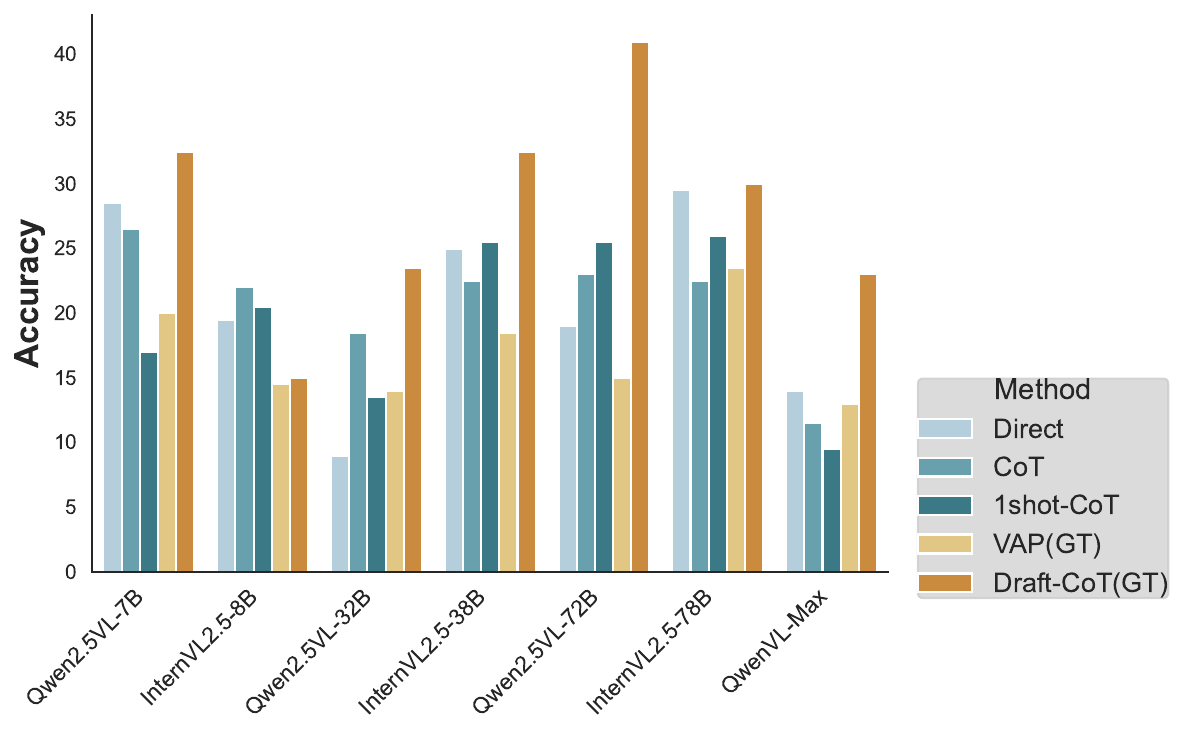}
    \caption{Accuracy with different models and methods in the hard Maze Judgment task. GT denotes that this result is obtained by ground truth in the route.}
    \label{fig:taska_hard}
\end{figure}

\paragraph{Limit gains of existing methods}
To investigate what factors can enhance the dynamic spatial reasoning capabilities of MLLMs, we conduct experiments on the hard judgment task using a variety of methods. As shown in Figure~\ref{fig:taska_hard}, various language-centric Chain-of-Thought approaches yield only marginal performance improvements and in some cases, even underperform compared to the original baseline. On the other hand, incorporating the VAP method with ground-truth positional images fails to improve model effectiveness and instead introduces noise that degrades performance. These results highlight the limitations of existing approaches and underscore the need for more effective integration of dynamic spatial information during the reasoning process.

\begin{figure}[t]
    \centering    \includegraphics[width=.8\linewidth]{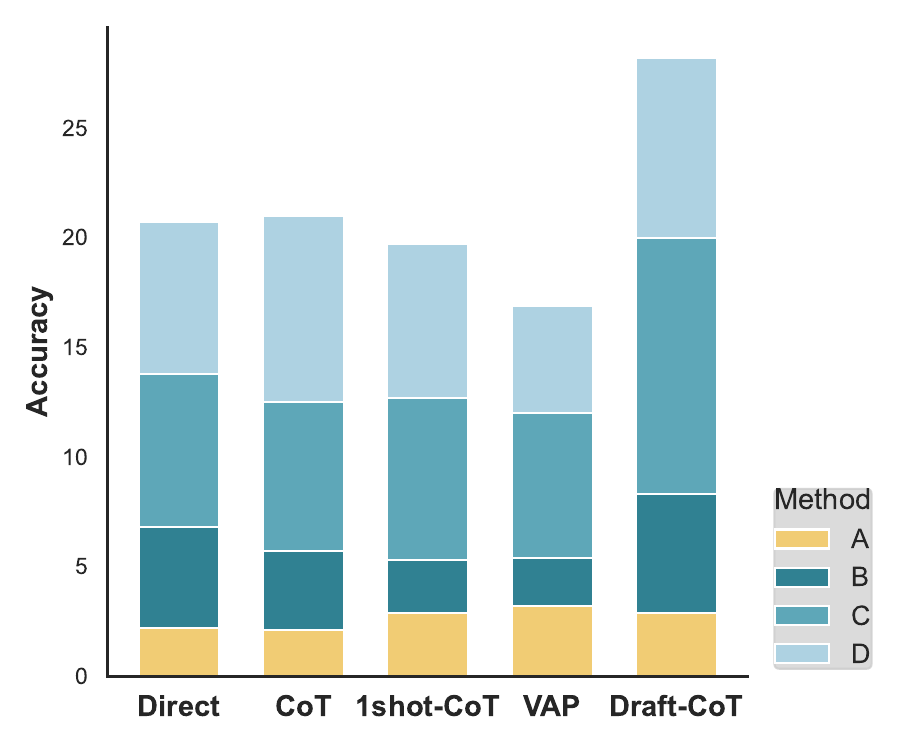}
    \caption{Average accuracy of models for each choice using various methods in the Maze Judgment task. }
    \label{fig:taska2}
\end{figure}

\paragraph{Drafts over dynamic images: Bring Surprise}
Inspired by the previous findings, we introduced visual navigation cues into the dynamic input images and combined them with textual CoT. This method allows the reasoning process to unfold through textual thought with its drafts over dynamic input images, termed as Draft CoT. Specifically, we directly edited the dynamic images by overlaying visual guidelines to depict the path. As shown in Figure~\ref{fig:taska_hard}, this approach significantly improves accuracy across all models, regardless of their underlying reasoning abilities, even outperforming the one-shot CoT setting in average accuracy. Moreover, as shown in Figure~\ref{fig:taska2}, the accuracy of all four options improves, rather than just increasing the success rate of a single option, further highlighting the robustness of Draft CoT across all scenarios. These results demonstrate the effectiveness of incorporating corresponding drafts over dynamic input images into the textual CoT process, providing new insights for dynamic spatial reasoning tasks.

\section{Methodology}
\begin{figure*}[h]
    \centering
    \includegraphics[width=\textwidth]{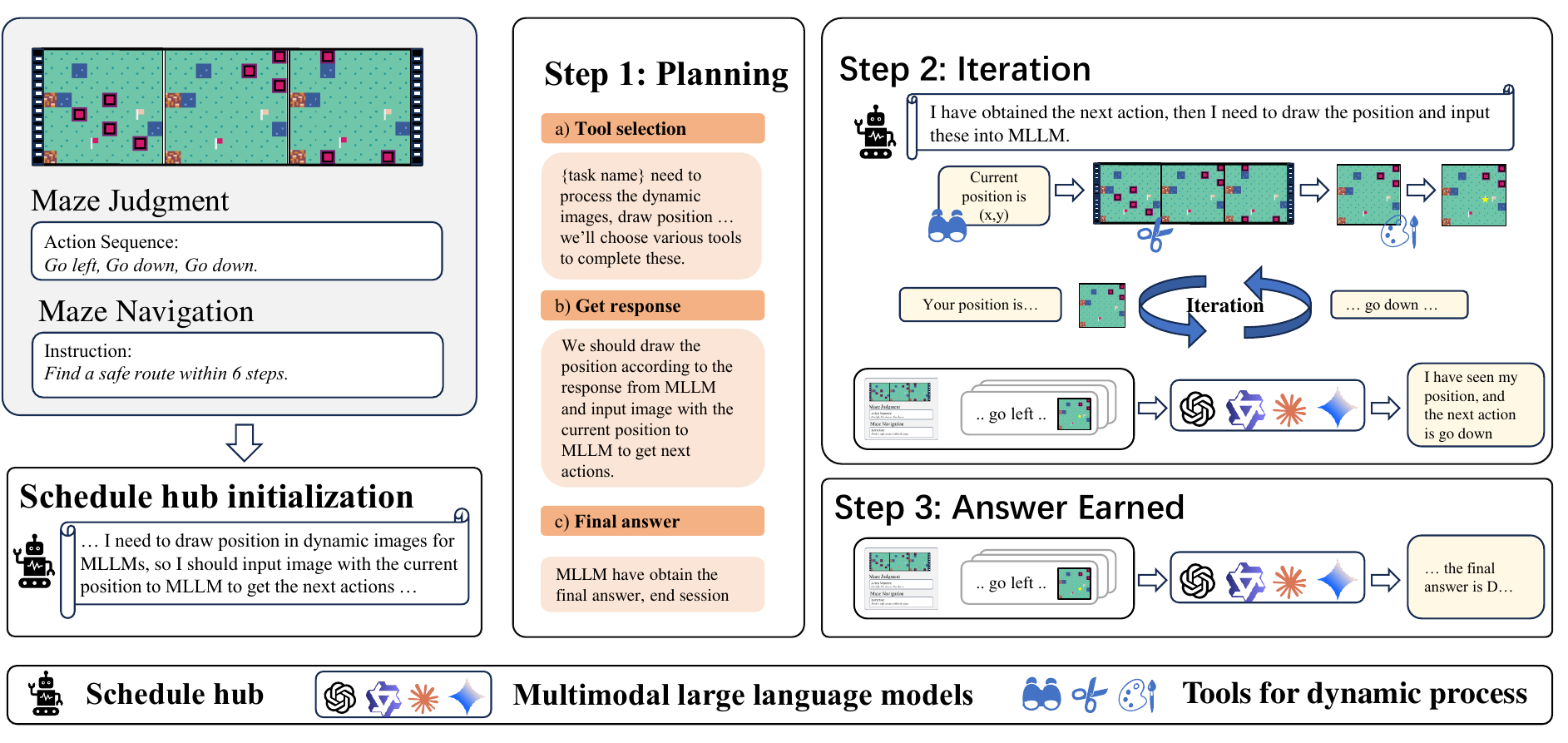} 
    \caption{Illustration of D2R reasoning process. After the schedule hub initialization, the process consists of planning, iteration, and answering three parts. }
    \label{fig:method}
\end{figure*}

Although the Draft CoT can obtain great performance gains, it rely on image generation capabilities not universally available across all the MLLMs.
To broaden its applications, we propose the Dynamic Draft Augmented Reasoning Framework (D2R), a training-free framework to generate intermediate thoughts on both textual thoughts and visual drafts. D2R extends the reasoning space from a signal language domain  $\mathcal{L}$ to multiple domains  $\mathcal{L} \cup \mathcal{V} \cup \mathcal{T}$, where $\mathcal{V}$ represents the visual domain and $\mathcal{T}$ represents the chronology domain. It enables models to reason in dynamic visual information by splitting it into steps and marking drafts over the input images in each step. By combining textual thoughts with corresponding drafts, this novel reasoning paradigm offers a more intuitive and accurate method with enhanced ability to collaborate on details between these two modalities.

\subsection{Toolkits for Synthesis and Drafting}
Drafting in the visual domain can enhance the ability to reason. However, MLLMs lack the ability to edit dynamic visual information and are weak in long text processing scheduling. Therefore, it is necessary to leverage external toolkits to enhance MLLM's performance. Therefore, we introduce the Dynamic-Information-Extract and Position-Draw tools for visual editing. Additionally, we also introduced an external LLM as a scheduling hub to organize the utilization of those tools.

\begin{table*}[t]
    \renewcommand{\arraystretch}{1.0}
    \centering
    \fontsize{11.6}{11.6}
    \selectfont
    \begin{tabular}{l|lcccc}
    \toprule
         \multirow{2}{*}{Model} & \multirow{2}{*}{Method} & \multicolumn{3}{c}{Total Acc}& \multirow{2}{*}{Average Acc}\\ 
         \cline{3-5} & & Easy & Normal & Hard\\ \midrule
         \multicolumn{6}{c}{\cellcolor[HTML]{F2F2F2}\textit{Maze Judgment task}} \\ \midrule
         \multirow{5}{*}{Qwen2.5VL-7B}
        ~ & Direct & \underline{22.5}   & \underline{34.0}   & \textbf{28.5}   & \underline{28.3}   \\ 
        ~ & CoT & 18.0(\textcolor{myred}{-4.5})   & 29.0(\textcolor{myred}{-5.0})   & 26.5(\textcolor{myred}{-2.0})   & 24.5(\textcolor{myred}{-3.8})   \\ 
        ~ & 1-shot CoT & 18.0(\textcolor{myred}{-4.5})   & 20.5(\textcolor{myred}{-13.5})   & 17.0(\textcolor{myred}{-11.5})   & 18.5(\textcolor{myred}{-9.8})   \\ 
        ~ & VAP & 13.5(\textcolor{myred}{-9.0})   & 15.0(\textcolor{myred}{-19.0})   & 20.0(\textcolor{myred}{-8.5})   & 16.2(\textcolor{myred}{-12.1})   \\ 
        ~ & \ourmethod (ours) & \textbf{34.0}(\textcolor{myblue}{+11.5})  & \textbf{46.0}(\textcolor{myblue}{+12.0})  & \underline{28.0}(\textcolor{myred}{-0.5})  & \textbf{36.0}(\textcolor{myblue}{+7.7})  \\ \midrule
        \multirow{5}{*}{Qwen2.5VL-72B}
        ~ & Direct & 61.0   & 38.5   & 19.0   & 39.5   \\ 
        ~ & CoT & \underline{67.0}(\textcolor{myblue}{+6.0})   & 40.0(\textcolor{myblue}{+1.5})   & 23.0(\textcolor{myblue}{+4.0})   & 43.3(\textcolor{myblue}{+3.8})   \\ 
        ~ & 1-shot CoT & \textbf{71.0}(\textcolor{myblue}{+10.0})   & \underline{46.5}(\textcolor{myblue}{+8.0})   & \underline{25.5}(\textcolor{myblue}{+6.5})   & \underline{47.7}(\textcolor{myblue}{+8.2})   \\ 
        ~ & VAP & 15.5(\textcolor{myred}{-45.5})   & 20.0(\textcolor{myred}{-18.5})   & 15.0(\textcolor{myred}{-4.0})   & 16.8(\textcolor{myred}{-22.7})   \\ 
        ~ & \ourmethod (ours) & \underline{67.0}(\textcolor{myblue}{+6.0})  & \textbf{49.0}(\textcolor{myblue}{+10.5})  & \textbf{41.0}(\textcolor{myblue}{+22.0})  & \textbf{52.3}(\textcolor{myblue}{+12.8})  \\ \midrule
        \multirow{5}{*}{QwenVL-max}
        ~ & Direct & \underline{40.0}   & 21.5   & \underline{14.0}   & \underline{25.2}   \\ 
        ~ & CoT & 36.0(\textcolor{myred}{-4.0})   & \underline{24.0}(\textcolor{myblue}{+2.5})   & 11.5(\textcolor{myred}{-2.5})   & 23.8(\textcolor{myred}{-1.4})   \\ 
        ~ & 1-shot CoT & 18.0(\textcolor{myred}{-22.0})   & 17.0(\textcolor{myred}{-4.5})   & 9.5(\textcolor{myred}{-4.5})   & 14.8(\textcolor{myred}{-10.4})   \\ 
        ~ & VAP & 15.0(\textcolor{myred}{-25.0})   & 9.0(\textcolor{myred}{-12.5})   & 13.0(\textcolor{myred}{-1.0})   & 12.3(\textcolor{myred}{-12.9})   \\ 
        ~ & \ourmethod (ours) & \textbf{46.5}(\textcolor{myblue}{+6.5})  & \textbf{35.5}(\textcolor{myblue}{+14.0})  & \textbf{28.0}(\textcolor{myblue}{+14.0})  & \textbf{36.7}(\textcolor{myblue}{+11.5})  \\ \midrule
        \multicolumn{6}{c}{\cellcolor[HTML]{F2F2F2}\textit{Maze Navigation task}} \\ \midrule
        \multirow{5}{*}{Qwen2.5VL-7B}
        ~ & Direct & 1.0 & \underline{2.0} & 1.0 & 1.3 \\ 
        ~ & CoT & 1.5(\textcolor{myblue}{+0.5}) & 1.5(\textcolor{myred}{-0.5}) & 0.0(\textcolor{myred}{-1.0}) & 1.0(\textcolor{myred}{-0.3}) \\ 
        ~ & 1-shot CoT & \underline{2.5}(\textcolor{myblue}{+1.5}) & \textbf{4.5}(\textcolor{myblue}{+2.5}) & \textbf{2.5}(\textcolor{myblue}{+1.5}) & \underline{3.2}(\textcolor{myblue}{+1.9}) \\
        ~ & VAP(GT)&-&-&-&-\\
        ~ & \ourmethod (ours) & \textbf{4.0}(\textcolor{myblue}{+3.0}) & \textbf{4.5}(\textcolor{myblue}{+2.5}) & \underline{2.0}(\textcolor{myblue}{+1.0}) & \textbf{3.5}(\textcolor{myblue}{+2.2}) \\ \midrule
        \multirow{5}{*}{Qwen2.5VL-72B} 
        & Direct & \underline{31.0} & \underline{21.5} & \underline{6.5} & \underline{19.7}\\
        & CoT & 16.5(\textcolor{myred}{-14.5}) & 17.5(\textcolor{myred}{-4.0}) & 0.5(\textcolor{myred}{-6.0}) & 11.5(\textcolor{myred}{-5.2})\\
        & 1Shot-CoT & 17.5(\textcolor{myred}{-13.5}) & 5.0(\textcolor{myred}{-16.5}) & 1.5(\textcolor{myred}{-5.0}) & 8.0(\textcolor{myred}{-11.7})\\
        & VAP(GT)&-&-&-&-\\
        & \ourmethod(ours) & \textbf{38.0}(\textcolor{myblue}{+7.0}) & \textbf{26.0}(\textcolor{myblue}{+4.5}) & \textbf{12.5}(\textcolor{myblue}{+6.0}) & \textbf{25.5}(\textcolor{myblue}{+5.8})\\ \midrule
        \multirow{5}{*}{QwenVL-max} 
        & Direct & 19.5 & \underline{10.5} & 1.5 & 10.5\\
        & CoT & \underline{22.5}(\textcolor{myblue}{+3.0}) & \underline{10.5} (-) & \underline{6.0}(\textcolor{myblue}{+4.5}) & \underline{13.0}(\textcolor{myblue}{+2.5})\\
        & 1Shot-CoT & 1.0(\textcolor{myred}{-18.5}) & 0.5(\textcolor{myred}{-10.0}) & 0.0(\textcolor{myred}{-1.5}) & 0.5(\textcolor{myred}{-10.0})\\
        & VAP(GT)&-&-&-&-\\
        & \ourmethod(ours) & \textbf{27.5}(\textcolor{myblue}{+8.0}) & \textbf{21.5}(\textcolor{myblue}{+11.0}) & \textbf{7.0}(\textcolor{myblue}{+5.5}) & \textbf{18.7}(\textcolor{myblue}{+8.2})\\
        \bottomrule
    \end{tabular}
    \caption{Performance of Maze Judgment task and Maze Navigation task. The results in ‘(·)’ represent the delta performance compared to the performance with direct prompt in each task. The best results of each dimension are \textbf{bold} and the secondary results are \underline{underlined}.}
    \label{tab:main_result}
\end{table*}

\subsection{Procedures of D2R}
We analogize D2R's process to an iterative process. Scheduled by the scheduling hub, D2R will autonomously determine the task type and generate a tool invocation plan, and it will maintain a real-time updated draft chain that is continuously supplemented with the most up-to-date information during the iteration process until the answer is generated. The whole process are as follows:

\begin{algorithm}[h]
    \caption{Procedures of Dynamic Draft Augmented Reasoning Framework}
    \KwIn{Text instruction $\mathbb{G}$ \\ \qquad \quad Dynamic images $\mathcal{I}$}
    \KwOut{Final answer $\mathcal{A}$}
    
    \textbf{Initialization:} \\
    $\mathcal{D}_p \gets$ Scheduling hub \\
    $\mathbb{E} \gets$ Tool set \\
    $\mathcal{C}_0 \gets \varnothing$, $n \gets 0$ \\
    
    \textbf{Step 1: Planning} \\
    $\varphi \gets \mathcal{D}_p(\mathbb{G}, \mathbb{E})$
    
    \textbf{Step 2: Iteration} \\
    \While{not $\mathcal{D}_p$ decides to stop\\} 
    {
        $c_{n} \gets \rm{MLLM}(\mathbb{G}, \mathcal{I}, \mathcal{C}_{< n})$ \\
        $e_n \gets \mathcal{D}_p(c_{n},\varphi)$\\
        $C_{n} \gets e_n (\varphi, \mathcal{I}, c_{n})$ \\
        $n \gets n + 1$
    }
    
    \textbf{Step 3: Final Answer} \\
    $\mathcal{A} \gets \rm{MLLM}(\mathbb{G}, \mathcal{I}, \mathcal{C}_{all})$

\end{algorithm}


\paragraph{Step 1: Planning} Our method takes a textual instruction $\mathbb{G}$ and dynamic images $\mathcal{I}$ as input. First, we prompt the scheduling hub to schedule a plan $\varphi$ and select the correct tools $e_n$ from the tool set $\mathbb{E}$. This step can be formalized as shown in Equation~\ref{eq1}:
\begin{equation}\label{eq1}
    \varphi \gets \mathcal{D}_p (\mathbb{G}, \mathbb{E}), 
\end{equation}
where $\mathcal{D}_p$ denotes the scheduling hub in this step.


\paragraph{Step 2: Iterative} As shown in Figure~\ref{fig:method}, after completing planning, D2R invokes the tool $e_n$ to generate the corresponding thought markers in images as drafts and fuse textual thought $c_n$ as augmented perceptual thought $\mathcal{C}_n$. In each iteration, $\mathcal{C}_n$ will be updated as the instruction progresses. The process is formally depicted as follows:
\begin{equation}\label{eq2}
    \left\{\begin{matrix} 
     c_{n} \gets \rm{MLLM}(\mathbb{G}, \mathcal{I}, \mathcal{C}_{< n})\\
     e_n \gets \mathcal{D}_p(c_{n},\varphi)\\
     C_{n} \gets e_n (\varphi, \mathcal{I}, c_{n})
    \end{matrix}\right. 
\end{equation}
where $\mathcal{C}_{< n}$ denotes the set of all the augmented perceptual thoughts before $n$ turns.

\paragraph{Step 3: Final Answer Iteration} When the iteration ends, scheduling hub will check the last output $c_{last}$ and determine if the answer $\mathcal{A}$ was generated. If $\mathcal{A}$ was not generated, scheduling hub will repeat the process and change the prompt strategy to instruct MLLM to output the answer As shown in Equation~\ref{eq3}, we take the set of all CoT $\mathcal{C}_{all}$ as input and use the prompt in the appendix to arrive at the final answer $\mathcal{A}$.
\begin{equation}\label{eq3}
    \mathcal{A} \gets \rm{MLLM}(\mathbb{G}, \mathcal{I}, \mathcal{C}_{all})
\end{equation}

\section{Experiment}

\begin{table*}[h]
    \centering
    \renewcommand{\arraystretch}{0.99}
    \begin{tabular}{lcccc}
    \toprule
         \multirow{2}{*}{Method} & \multicolumn{3}{c}{Total Acc}& \multirow{2}{*}{Average Acc}\\ 
         \cline{2-4} & Easy & Normal & Hard\\ \midrule
        Qwen2.5VL-72B(D2R) & 67.0 & 49.0 & 41.0 &52.3\\ 
        w/o Textual Thought & 52.0(\textcolor{myred}{-15.0}) & 44.3(\textcolor{myred}{-4.7}) & 32.0(\textcolor{myred}{-9.0}) & 42.7(\textcolor{myred}{-9.6})\\ 
        w/o Drafts over dynamic images & 45.1(\textcolor{myred}{-21.9}) & 33.3(\textcolor{myred}{-15.7}) & 17.1(\textcolor{myred}{-23.9}) &31.8(\textcolor{myred}{-20.5})\\ \bottomrule
    \end{tabular}
    \caption{The accuracy of the removal of drafts over dynamic images or textual thought in \ourmethod of the maze judgment task. The results in ‘(·)’ represent the delta performance compared to \ourmethod with both two modalities.}
    \label{tab:ablation}
    \vspace{4pt}
\end{table*}

\begin{table}[]
    \renewcommand{\arraystretch}{1.05}
    \selectfont
    \centering
    \resizebox{.41\textwidth}{!}{%
    \begin{tabular}{llc}
    \toprule
        Model & Method & Acc \\ \midrule
        \multirow{2}{*}{Qwen2.5VL-7B} 
        & Draft CoT(GT) & 32.5 \\
        & \ourmethod & 28.0  \\
        \hline
        \multirow{2}{*}{Qwen2.5VL-72B} 
        & Draft CoT(GT) & 41.0\\
        & \ourmethod & 41.0\\
        \hline
        \multirow{2}{*}{QwenVL-max} 
        & Draft CoT(GT) & 22.0\\
        & \ourmethod & 28.0 \\
        \bottomrule
    \end{tabular}}
        \caption{Performance of hard maze judgment between D2-CoT(GT) and D2R among three models.}
    \label{tab:vsGT}
    \vspace{-7pt}
\end{table}

\subsection{Experiment Setup}
We construct datasets for two dynamic spatial reasoning tasks described in Section 3, encompassing three levels of complexity in environment and action spaces. We use Qwen-Max as the scheduling hub in our work, and the temperature is set to 0.1. We compare the D2R with the following reasoning methods: 1) Direct Prompt. 2)Chain-of-thought (CoT). 3) CoT with 1-shot. 4)VAP. In our experiments, we use Qwen2.5-VL-7B, Qwen2.5-VL-72B, and Qwen-VL-Max as the MLLM part of D2R.

\subsection{D2R has better dynamic reasoning ability}

As shown in Table~\ref{tab:main_result}, both two tasks show that D2R demonstrates greater stability and accuracy.
In the maze judgment task,  direct and language-centric CoT methods perform comparably to D2R under low-difficulty conditions, their accuracy declines significantly as task complexity increases. This suggests that textual Chain-of-Thought reasoning is insufficient for handling more complex scenarios. In contrast, the performance gap widens in favor of D2R as difficulty increases, highlighting its robustness and effectiveness under challenging conditions. Furthermore, D2R also shows higher accuracy in the maze navigation task. It is important to note that our method achieves performance improvements across all models and difficulties. This underscores the crucial role of integrating both textual thought and their drafts in dynamic planning tasks, as such collaboration enhances the model's ability to effectively handle complex reasoning scenarios.

\subsection{How D2R is effective?}

\paragraph{Can D2R be effective with different MLLMs' abilities? }
To further explore the effectiveness of our method with different models' ability, we conduct experiments on three MLLMs and the results are shown in Table~\ref{tab:main_result}.
Although the effect varies with basic model ability and task difficulty, we can still enhance the capabilities of different models: all three MLLMs can perform better than the basics in most cases. However, Qwen2.5-VL-72B and QwenVL-max gain substantially more from D2R than Qwen2.5-VL-7B, highlighting the challenges faced by less capable models in fully utilizing our method. In other words, while D2R can help externalize the reasoning process of MLLM, it cannot fundamentally improve the inherent reasoning capacity of the model.

\paragraph{Are drafts and texts equally important?}
To further validate the contribution of the textual thought and drafts over dynamic images to D2R, we experiment by removing textual thoughts and corresponding drafts in the maze judgment task, respectively. As shown in Table~\ref{tab:ablation}, the removal of any component from either part leads to a performance decline across all difficulty levels, reflecting the importance of integrating both textual thought and its drafts in reasoning. Notably, performance drops more significantly when the drafts are removed than when the textual thoughts are removed, further proving the crucial role of draft processing in dynamic spatial reasoning.

\paragraph{Can D2R be as effective as Draft CoT(GT)?}
To explore whether our methods can reach the same performance with draft DoT(GT), we compare the experimental results between the D2R and Draft CoT(GT). As shown in Table~\ref{tab:vsGT}, compared to the results with Draft CoT(GT), all three models can obtain comparable performance using our methods.
The results show that our method can successfully make the MLLMs detect the current position and output the next action to accomplish different tasks in most cases, resulting in only a small gap from the ground truth.

\section{Conclusion}

In this paper, we introduce GRASSLAND and present two tasks to evaluate the performance on dynamic multimodal spatial reasoning: Maze Judgment and Maze Navigation. Through experiments, we observe that the combination of the textual thoughts and their drafts over dynamic input images, termed Draft CoT, significantly outperforms other approaches in these tasks, providing new insights into the dynamic spatial reasoning process. To make Draft CoT more widely applicable in existing MLLMs, we propose the Dynamic Draft Augmented Reasoning Framework, a training-free framework that generates intermediate thoughts by combining both textual thoughts and their drafts over dynamic input images. Experimental results show that D2R delivers exceptional performance across various dynamic spatial reasoning tasks.

\section*{Limitation}
While D2R significantly outperforms other methods that do not require training under multiple tasks, the performance gains are different among various models, especially the weaker models gain less than the stronger models. This discrepancy suggests that D2R's benefits are more pronounced in models with a higher baseline capacity, highlighting its potential to enhance the performance of more powerful architectures more effectively. Moving forward, we plan to explore strategies for improving D2R's applicability to weaker models, aiming to achieve more excellent performance across a broader range of architectures.

\section*{Ethic Consideration}
Our data is generated through open-source software and our own proprietary code. All models used are open-source, and their sources are clearly credited. The entire process follows transparent and ethical guidelines, ensuring there are no ethical concerns or issues with the data generation. We are committed to maintaining high standards of integrity and transparency in our work.

\clearpage

\bibliography{appendix/acl_latex}
\appendix
\clearpage

\section{Details on MLLMs}
Table \ref{app:hyperparam} shows the hyperparameters for generating with MLLM and size information for each model. For QwenVL-Max and Qwen-Max, we use the 2025-01-25 version through Aliyun platform.
\begin{table}[h]
    \centering
    \begin{tabular}{lccc}
    \toprule
        Model  & Max tokens & Size \\ \midrule
        QwenVL2.5  & 700 & 72B, 32B, 7B \\ 
        InternVL2.5 & 700 & 78B, 38B, 8B \\ 
        QwenVL-Max* & 700 & -  \\ 
        VideoLLaMA3  & 700 & 7B \\ 
        Qwen-Max*  & 400 & - \\ \bottomrule
    \end{tabular}
    \caption{Hyperparameters for model generation. Model called via API has been marked by *}
    \label{app:hyperparam}
\end{table}

\section{Metric}
We use the accuracy as the evaluation metric for both two tasks. For the maze judgment task, the accuracy aims to detect whether the model can obtain the final state. For the maze navigation task, the accuracy aims to detect whether the model can reach the final position according to the model's response. 




\section{Other results}
The other results about our methods are presented in Table~\ref{app:taska} and Table~\ref{app:taskb}. Specifically,
Table~\ref{app:taska} and Table~\ref{app:taskb} presents the detailed performance across various methods for each task. For maze judgment task, we observe a clear uneven distribution of answer accuracies on other methods, with answer "D. Action Failed: Agent Safe but Fail to Reach Destination" being significantly more accurate than the other three options. It reflects the shortcomings of inadequate ability to judge complex states on these methods. In contrast, D2R outperforms other methods, optimizing accuracy on the complex options A and B.

For the maze navigation task, we notice an interesting feature in all methods that in the correct path answer, the effective length is shorter than the full path length. It means the goal point is reached at the halfway. Even D2R can only make the gap smaller, not eliminate it completely. This reflects a possible deficiency in the model to perform spatial planning tasks.

\begin{table*}
    \centering
    \fontsize{10.5}{9.5}
    \selectfont
    \begin{tabular}{llccccc}
    \toprule
      \multirow{2}{*}{Model} & \multirow{2}{*}{Method} & \multicolumn{4}{c}{Choice Acc.} & \multirow{2}{*}{Total Acc.}  \\
      \cline{3-6} & & A & B & C & D\\ \midrule
       \multicolumn{7}{c}{\cellcolor[HTML]{F2F2F2}\textit{Easy Level}} \\ \midrule
        \multirow{5}{*}{Qwen2.5VL-7B}
        ~ & Direct & 45.0  & - & 13.2  & 21.8  & 22.5   \\ 
        ~ & CoT & 35.0  & - & 23.3  & 13.3  & 18.0   \\ 
        ~ & 1-shot CoT & 75.0  & - & 15.7  & 10.5  & 18.0   \\ 
        ~ & VAP & 75.0  & - & 18.4  & 3.5  & 13.5   \\ 
        ~ & D2R (ours) & 65.0  & - & 23.7  & 32.4  & 34.0  \\ \midrule
        \multirow{5}{*}{Qwen2.5VL-72B}
        ~ & Direct & 25.0  & - & 15.8  & 78.2  & 61.0   \\ 
        ~ & CoT & 40.0  & - & 10.5  & 85.9  & 67.0   \\ 
        ~ & 1-shot CoT & 10.0  & - & 5.3  & 97.2  & 71.0   \\ 
        ~ & VAP & 10.0  & - & 5.3  & 19.0  & 15.5   \\ 
        ~ & D2R (ours) & 30.0  & - & 21.1  & 84.5  & 67.0  \\ \midrule
        \multirow{5}{*}{QwenVL-Max}
        ~ & Direct & 35.0  & - & 10.5  & 48.6  & 40.0   \\ 
        ~ & CoT & 35.0  & - & 10.5  & 43.0  & 36.0   \\ 
        ~ & 1-shot CoT & 30.0  & - & 7.9  & 19.0  & 18.0   \\ 
        ~ & VAP & 35.0  & - & 5.3  & 14.8  & 15.0   \\ 
        ~ & D2R (ours) & 40.0  & - & 2.6  & 59.2  & 46.5  \\ \midrule
        \multicolumn{7}{c}{\cellcolor[HTML]{F2F2F2}\textit{Normal Level}} \\ \midrule
        \multirow{5}{*}{Qwen2.5VL-7B}
        ~ & Direct & 46.7  & 325.3  & 22.2  & 40.6  & 34.0   \\ 
        ~ & CoT & 46.7  & 23.5  & 25.0  & 30.2  & 29.0   \\ 
        ~ & 1-shot CoT & 13.3  & 5.9  & 5.6  & 26.0  & 20.5   \\ 
        ~ & VAP & 80.0  & 0.0  & 8.3  & 12.5  & 15.0   \\ 
        ~ & D2R (ours) & 33.3  & 0.0  & 9.7  & 83.3  & 46.0  \\ \midrule
        \multirow{5}{*}{Qwen2.5VL-72B}
        ~ & Direct & 6.7  & 17.6  & 11.1  & 67.7  & 38.5   \\ 
        ~ & CoT & 6.7  & 11.8  & 8.3  & 74.0  & 40.0   \\ 
        ~ & 1-shot CoT & 13.3  & 5.9  & 6.9  & 88.5  & 46.5   \\ 
        ~ & VAP & 20.0  & 0.0  & 4.2  & 35.4  & 20.0   \\ 
        ~ & D2R (ours) & 33.3  & 47.1  & 12.5  & 79.2  & 49.0  \\ \midrule
        \multirow{5}{*}{QwenVL-Max}
        ~ & Direct & 26.7  & 0.0  & 8.3  & 34.4  & 21.5   \\ 
        ~ & CoT & 40.0  & 5.9  & 4.2  & 39.6  & 24.0   \\ 
        ~ & 1-shot CoT & 52.6  & 20.8  & 4.9  & 19.1  & 17.0   \\ 
        ~ & VAP & 13.3  & 0.0  & 11.1  & 8.3  & 9.0   \\ 
        ~ & D2R (ours) & 26.7  & 11.8  & 2.8  & 65.6  & 35.5  \\ \midrule
        \multicolumn{7}{c}{\cellcolor[HTML]{F2F2F2}\textit{Hard Level}} \\ \midrule
        \multirow{5}{*}{Qwen2.5VL-7B}
        ~ & Direct & 21.1  & 54.7  & 8.6  & 36.2  & 28.5   \\ 
        ~ & CoT & 15.8  & 43.4  & 18.5  & 25.5  & 26.5   \\ 
        ~ & 1-shot CoT & 52.6  & 20.8  & 4.9  & 19.1  & 17.0   \\ 
        ~ & VAP & 89.5  & 7.5  & 13.6  & 17.0  & 20.0   \\ 
        ~ & D2R (ours) & 15.8  & 1.9  & 11.0  & 91.0  & 28.0  \\ \midrule
        \multirow{5}{*}{Qwen2.5VL-72B}
        ~ & Direct & 10.5  & 9.4  & 4.9  & 57.4  & 19.0   \\ 
        ~ & CoT & 10.5  & 5.7  & 9.9  & 70.2  & 23.0   \\ 
        ~ & 1-shot CoT & 5.3  & 0.0  & 8.6  & 91.5  & 25.5   \\ 
        ~ & VAP & 10.5  & 0.0  & 1.2  & 57.4  & 15.0   \\ 
        ~ & D2R (ours) & 15.8  & 39.6  & 19.8  & 89.4  & 41.0  \\ \midrule
        \multirow{5}{*}{QwenVL-Max}
        ~ & Direct & 31.6  & 1.9  & 9.9  & 27.7  & 14.0   \\ 
        ~ & CoT & 42.1  & 0.0  & 6.2  & 21.3  & 11.5   \\ 
        ~ & 1-shot CoT & 21.1  & 7.5  & 8.6  & 8.5  & 9.5   \\ 
        ~ & VAP & 42.1  & 0.0  & 14.8  & 12.8  & 13.0   \\ 
        ~ & D2R (ours) & 21.1  & 20.8  & 6.2  & 76.6  & 28.0  \\ 
        \bottomrule
    \end{tabular}
    \caption{Detailed performance on maze judgment task.}
    \label{app:taska}
\end{table*}

\begin{table*}
    \centering
    \fontsize{10.5}{9.5}
    \selectfont
    \begin{tabular}{llccccc}
    \toprule
        Model & Method & Arrived & Failed & Unfinished & \makecell{Ave. Step \\ (Effective)} & \makecell{Ave. Step \\ (Answer)}  \\ \midrule
       \multicolumn{7}{c}{\cellcolor[HTML]{F2F2F2}\textit{Easy Level}} \\ \midrule
       \multirow{4}{*}{Qwem2.5VL-7B}
        ~ & Direct & 1.0 & 4.0 & 95.0 & 4.50  & 6.00   \\ 
        ~ & CoT & 1.5 & 9.5 & 89.0 & 4.00  & 5.67   \\ 
        ~ & 1-shot CoT & 2.5 & 62.0 & 35.5 & 4.40  & 5.80   \\ 
        ~ & D2R (ours) & 4.0 & 38.0 & 58.0 & 3.88  & 6.00  \\  \midrule
       \multirow{4}{*}{Qwen2.5VL-72B}
        & Direct & 31.0 & 40.0 & 29.0 & 3.55 & 5.84  \\ 
        & CoT & 16.5 & 43.5 & 40.0 & 3.48 & 5.97  \\ 
        & 1-shot CoT & 17.5 & 35.5 & 47.0 & 3.54 & 5.91  \\ 
        & D2R (ours) & 38.0 & 38.0 & 24.0 & 3.72 & 5.74 \\  \midrule
        \multirow{4}{*}{QwenVL-Max}
        & Direct & 19.5 & 30.5 & 50.0 & 3.31 & 5.97  \\ 
        & CoT & 22.5 & 39.5 & 38.0 & 3.56 & 5.93  \\ 
        & 1-shot CoT & 1.0 & 41.0 & 58.0 & 6.00 & 6.00  \\ 
        & D2R (ours) & 27.5 & 41.0 & 31.5 & 4.04 & 5.47 \\ \midrule
        \multicolumn{7}{c}{\cellcolor[HTML]{F2F2F2}\textit{Normal Level}} \\ \midrule
        \multirow{4}{*}{Qwem2.5VL-7B}
        ~ & Direct & 2.0 & 8.5 & 89.5 & 4.00  & 5.75   \\ 
        ~ & CoT & 1.5 & 13.5 & 85.0 & 3.67  & 6.00   \\ 
        ~ & 1-shot CoT & 4.5 & 52.5 & 43.0 & 3.78  & 5.56   \\ 
        ~ & D2R (ours) & 4.5 & 52.5 & 43.0 & 4.67  & 6.00 \\ \midrule
        \multirow{4}{*}{Qwen2.5VL-72B}
        & Direct & 21.5 & 58.5 & 20.0 & 3.72 & 5.77  \\ 
        & CoT & 17.5 & 48.5 & 34.0 & 3.89 & 6.06  \\ 
        & 1shot-CoT & 5.0 & 56.5 & 38.5 & 3.50 & 6.00  \\ 
        & D2R (ours) & 26.0 & 51.0 & 23.0 & 3.94 & 5.98 \\ \midrule
        \multirow{4}{*}{QwenVL-Max}
        & Direct & 10.5 & 40.0 & 49.5 & 3.52 & 6.00  \\ 
        & CoT & 10.5 & 38.0 & 51.5 & 3.71 & 6.05  \\ 
        & 1-shot CoT & 0.5 & 50.5 & 49.0 & 6.00 & 6.00  \\ 
        & D2R (ours) & 21.5 & 54.0 & 24.5 & 3.93 & 5.93 \\ \midrule
        \multicolumn{7}{c}{\cellcolor[HTML]{F2F2F2}\textit{Hard Level}} \\ \midrule
        \multirow{4}{*}{Qwem2.5VL-7B}
        ~ & Direct & 1.0 & 9.5 & 89.5 & 4.00  & 5.50   \\ 
        ~ & CoT & 0.0 & 12.0 & 88.0 & 0.00  & 0.00   \\ 
        ~ & 1-shot CoT & 2.5 & 62.0 & 35.5 & 4.40  & 5.80   \\ 
        ~ & D2R (ours) & 2.0 & 63.0 & 35.0 & 5.25  & 6.00  \\ \midrule
        \multirow{4}{*}{Qwen2.5VL-72B}
        & Direct & 6.5 & 74.5 & 19.0 & 4.69 & 5.69  \\ 
        & CoT & 0.5 & 72.0 & 27.5 & 4.00 & 6.00  \\ 
        & 1-shot CoT & 1.5 & 67.5 & 31.0 & 4.67 & 6.00  \\ 
        & D2R (ours) & 12.5 & 75.5 & 12.0 & 4.48 & 6.00 \\ \midrule
        \multirow{4}{*}{QwenVL-Max}
        & Direct & 1.5 & 61.0 & 37.5 & 4.67 & 6.00  \\ 
        & CoT & 6.0 & 59.5 & 34.5 & 4.17 & 6.00  \\ 
        & 1-shot CoT & 0.0 & 62.5 & 37.5 & 0.00 & 0.00  \\ 
        & D2R (ours) & 7.0 & 77.5 & 15.5 & 4.86 & 5.85 \\ 
        \bottomrule
    \end{tabular}
    \caption{Detailed performance on maze navigation task.}
    \label{app:taskb}
\end{table*}

\section{Prompt}
\subsection{Basic Prompt}
Table~\ref{app:direct} shows the prompting template of direct reasoning and D2R task prompt for each task. Table~\ref{app:cot} and Table~\ref{app:1cot} shows the prompt for each task with different reasoning methods. 
\begin{table*}
    \centering
    \begin{tcolorbox}[title={Direct}]
            Task:Maze Judgment
            
            Tile info: character can move pass the green tile(grass). The red tile(lava) and the blue tile(water) is dangerous, when player move into them, game fail. The yellow and brown tile is wall which can not be passed.
            
            The white flag is start point, the red flag is destination point
            
            Player can't move off the map, considering it as air walls
            
            Actions: the lava tile change position every second, and player also move every second. Consider player move first in same time, which mean if player and lava tile move to same position, the game fail.
            
            Determine whether the agent (elf character) can safely reach the destination following the action sequence without falling into the lava or water. If not, identify the failure reason shortly. The definitions of the actions are as below.
            
            * In the video, the red line shows the movement path of the agent.
            
            * Go up/left/down/right: move one grid space in the absolute up/left/down/right direction.
            
            After analyse the actions, return A, B, C or D.
            
            Full Action Sequence: {action\_sequence}
            
            A. Action Success.
            
            B. Action Failed: Fall into the water.
            
            C. Action Failed: Fall into the lava.
            
            D. Action Failed: Agent Safe but Fail to Reach Destination.
            \tcblower
            Task: Route Plan
            
            Tile Info: The character can move across the green tile (grass). The red tile (lava) and the blue tile (water) are dangerous. If the player moves onto them, the game fails. The yellow and brown tiles are walls, which cannot be passed. The white flag represents the starting point, and the red flag represents the destination.
            
            The player cannot move off the map; treat the edges as air walls.
            Actions: The lava tiles change position every second, and the player also moves every second.
            
            Consider the player moving first in the same time step, which means if the player and a lava tile move to the same position, the game fails.
            
            You will receive a 6-second video showing the dynamic map. Your task is to analyze this video, apply the rules mentioned above, then determine a route that allows the player to reach the destination safely within 6 steps.
            
            The answer should follow this format: "Action: [START] Go right, Go up, Go down, ... [END]" Each command corresponds to one move. And put it at the end of your answer.
            
            Move Commands: Go up/left/down/right: Move one grid space in the absolute up/left/down/right direction.
    \end{tcolorbox}
    
    \caption{Example of input for Direct reasoning}
    \label{app:direct}
\end{table*}

\begin{table*}
    \centering
    \begin{tcolorbox}[title={CoT reasoning}]
           Task:Maze Judgment
            
            Tile info: character can move pass the green tile(grass). The red tile(lava) and the blue tile(water) is dangerous, when player move into them, game fail. The yellow and brown tile is wall which can not be passed.
            
            The white flag is start point, the red flag is destination point
            
            Player can't move off the map, considering it as air walls
            
            Actions: the lava tile change position every second, and player also move every second. Consider player move first in same time, which mean if player and lava tile move to same position, the game fail.
            
            Determine whether the agent (elf character) can safely reach the destination following the action sequence without falling into the lava or water. If not, identify the failure reason shortly. The definitions of the actions are as below.
            
            * In the video, the red line shows the movement path of the agent.
            
            * Go up/left/down/right: move one grid space in the absolute up/left/down/right direction.
            
            After analyse the actions, return A, B, C or D.
            
            Full Action Sequence: {action\_sequence}
            
            A. Action Success.
            
            B. Action Failed: Fall into the water.
            
            C. Action Failed: Fall into the lava.
            
            D. Action Failed: Agent Safe but Fail to Reach Destination.

            Let's think it step-by-step and make right choice.
            \tcblower
            Task: Route Plan
            
            Tile Info: The character can move across the green tile (grass). The red tile (lava) and the blue tile (water) are dangerous. If the player moves onto them, the game fails. The yellow and brown tiles are walls, which cannot be passed. The white flag represents the starting point, and the red flag represents the destination.
            
            The player cannot move off the map; treat the edges as air walls.
            Actions: The lava tiles change position every second, and the player also moves every second.
            
            Consider the player moving first in the same time step, which means if the player and a lava tile move to the same position, the game fails.
            
            You will receive a 6-second video showing the dynamic map. Your task is to analyze this video, apply the rules mentioned above, then determine a route that allows the player to reach the destination safely within 6 steps.
            
            The answer should follow this format: "Action: [START] Go right, Go up, Go down, ... [END]" Each command corresponds to one move. And put it at the end of your answer.
            
            Move Commands: Go up/left/down/right: Move one grid space in the absolute up/left/down/right direction.

            Let's think it step-by-step and make right choice.
    \end{tcolorbox}
    \caption{Example of input for CoT reasoning}
    \label{app:cot}
\end{table*}

\begin{table*}
    \centering
    \begin{tcolorbox}[title={CoT with 1-shot prompting}]
           Task:Maze Judgment
            
            Tile info: character can move pass the green tile(grass). The red tile(lava) and the blue tile(water) is dangerous, when player move into them, game fail. The yellow and brown tile is wall which can not be passed.
            
            The white flag is start point, the red flag is destination point
            
            Player can't move off the map, considering it as air walls
            
            Actions: the lava tile change position every second, and player also move every second. Consider player move first in same time, which mean if player and lava tile move to same position, the game fail.
            
            Determine whether the agent (elf character) can safely reach the destination following the action sequence without falling into the lava or water. If not, identify the failure reason shortly. The definitions of the actions are as below.
            
            * In the video, the red line shows the movement path of the agent.
            
            * Go up/left/down/right: move one grid space in the absolute up/left/down/right direction.
            
            After analyse the actions, return A, B, C or D.
            
            Full Action Sequence: {action\_sequence}
            
            A. Action Success.
            
            B. Action Failed: Fall into the water.
            
            C. Action Failed: Fall into the lava.
            
            D. Action Failed: Agent Safe but Fail to Reach Destination.

            Here is an example, consider video follow behind the text. The action sequence is: Go down, Go up, Go up, Go left. First, the agent move down. Check the tile agent move to, it is grass with no trap, so agent can move to. Then agent move up, it is start point, agent can move to here. Then agent move up again, it is grass, agent can move to here. Then agent move left, it is the end point, so agent arrive at the destination. So the answer is: A. Action Success.

            Video: <example\_video>
            \tcblower
            Task: Route Plan
            
            Tile Info: The character can move across the green tile (grass). The red tile (lava) and the blue tile (water) are dangerous. If the player moves onto them, the game fails. The yellow and brown tiles are walls, which cannot be passed. The white flag represents the starting point, and the red flag represents the destination.
            
            The player cannot move off the map; treat the edges as air walls.
            Actions: The lava tiles change position every second, and the player also moves every second.
            
            Consider the player moving first in the same time step, which means if the player and a lava tile move to the same position, the game fails.
            
            You will receive a 6-second video showing the dynamic map. Your task is to analyze this video, apply the rules mentioned above, then determine a route that allows the player to reach the destination safely within 6 steps.
            
            The answer should follow this format: "Action: [START] Go right, Go up, Go down, ... [END]" Each command corresponds to one move. And put it at the end of your answer.
            
            Move Commands: Go up/left/down/right: Move one grid space in the absolute up/left/down/right direction.
            
            Here is an example, consider video follow behind the text. To move safely, we check the position of destination, make choice, and review the traps position in video to conform the action safe. In this example, the best action is: [START] Go right, Go right, Go right, Go right, Go right, Go down [END]

            Video: <example\_video>
    
    \end{tcolorbox}
    \caption{Example of input for CoT reasoning with 1-shot prompting}
    \label{app:1cot}
\end{table*}

\subsection{Method Prompt}
Table~\ref{app:plan} shows the example of prompt for scheduling hub. Table~\ref{app:modelrep} shows the prompt in iteration process for each task in D2R.
\begin{table*}[]
    \centering
    \begin{tcolorbox}[title={Planning prompt for manager}]
    You are controlling the VideoProcessing agent, PositionGet agent, DrawPosition agent and MLLMReply agent. 
    
    1.Each time you need to extract and save the video by VideoProcessing agent, get the postion by PositionGet agent, and draw the position by DrawPosition agent.
    
    2.you need to complete the task by MLLMReply agent until the MLLMReply agent output final answer.
    
    3.If the MLLMReply agent do not output the final answer, you need to continue completing the task by MLLMReply agent.
    
    4.You should follow the order in examples, and don't make any superfluous execution.
    
    5.When MLLMReply agent output <finish>, you need to finish the task.
    \end{tcolorbox}
    \caption{Example of input for manager LLM}
    \label{app:plan}
\end{table*}

\begin{table*}[]
    \centering
    \begin{tcolorbox}[title={Iteration prompt for MLLM}]
        Additionly, there are the <visualization of the thought>, you should output next action until the final answer is obtained. Here are some tips you should follow:
        
        1.<visualization of the thought> consist of the image of now position and the text of next action.
        
        2.The black square represents your current position.
        
        3.Based on the <Task Description> and <visualization of the thought>, follow the next action in <Full Action Sequence>. Each action in <visualization of the thought> has already been executed, don't execute them again.
        
        4.If the next action is wall or move off the map, append <can\_not\_pass> after the next action.
        
        5.As soon as you can get the final answer, you will immediately output the final answer and append <finish> after it.
            
        6.Each time you can only output one action or one final answer.
    \end{tcolorbox}
    \caption{Example of input system prompt for MLLM in iteration process}
    \label{app:modelrep}
\end{table*}
\section{Case Study}
\subsection{Maze Judgment}
\begin{figure*}[h]
    \centering
    \includegraphics[width=\textwidth]{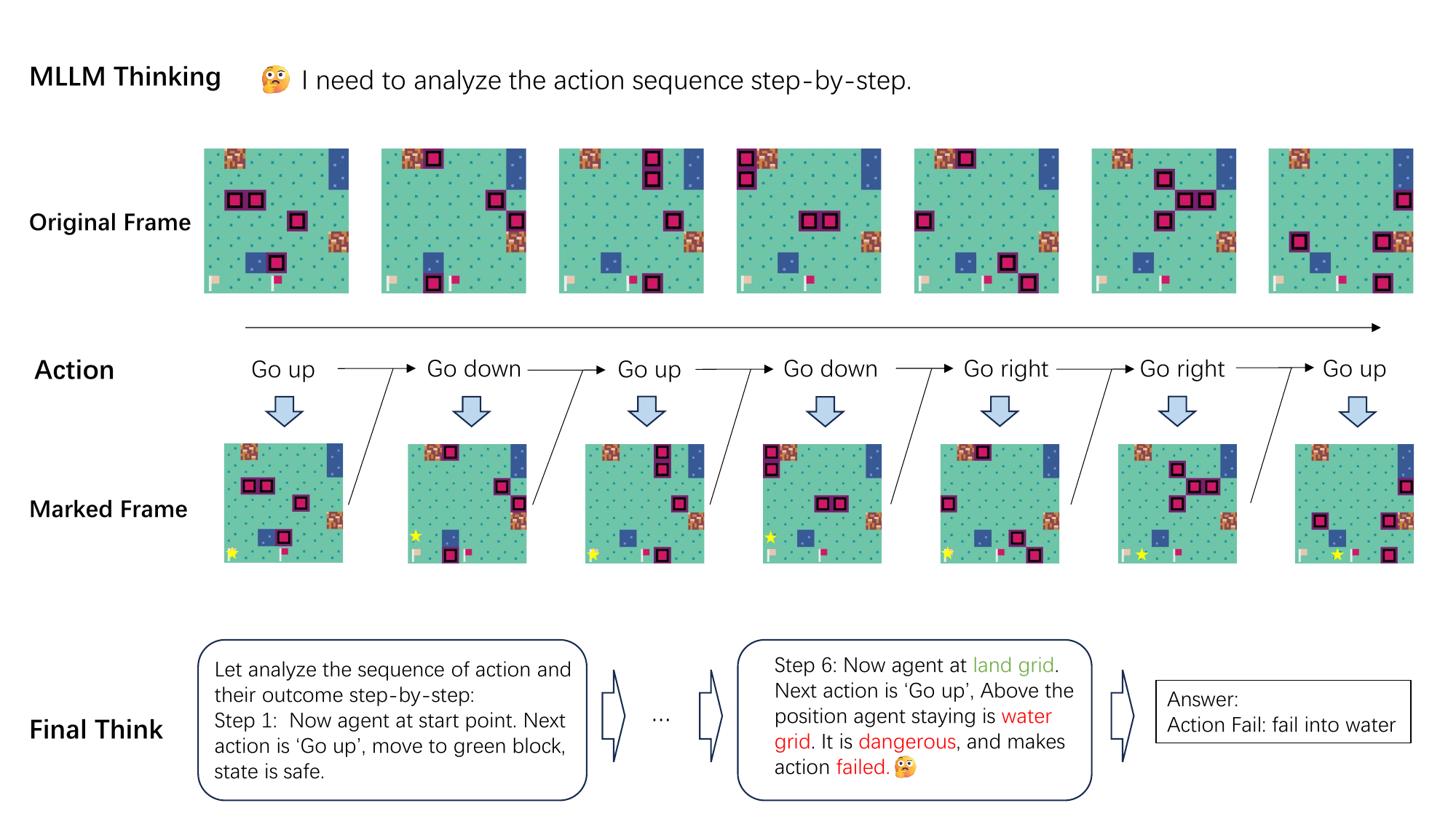} 
    \caption{An example of the thought process for D2R in maze judgment task.}
    \label{fig:case1}
\end{figure*}

Figure~\ref{fig:case1} presents the thought process of D2R in maze judgment task. In each step, after receiving the action instruction, D2R mark the original frame with the position staying now, then searches the grids in the action direction to judge the state after the action is executed.
\subsection{Maze Navigation}
\begin{figure*}[h]
    \centering
    \includegraphics[width=\textwidth]{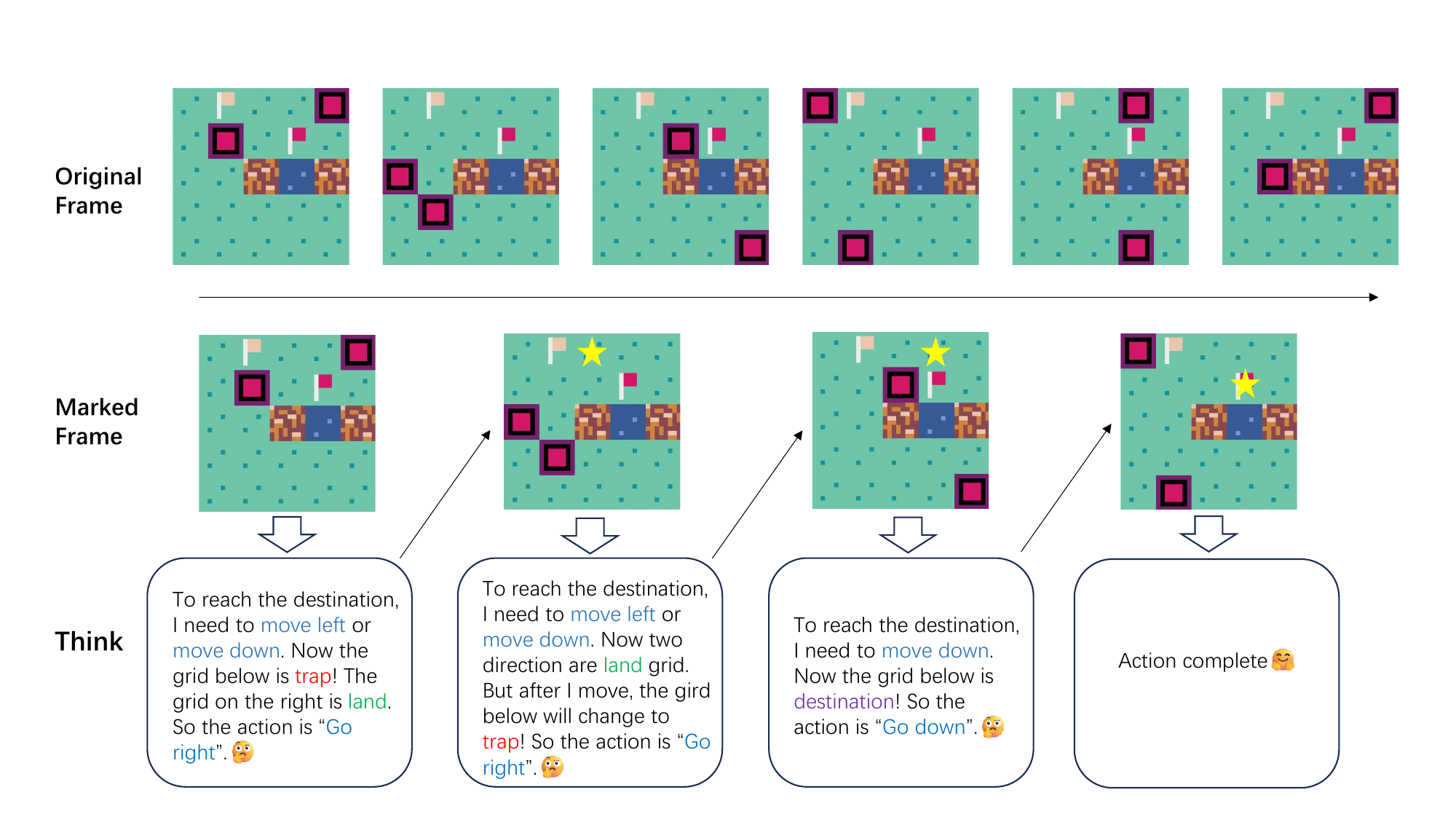} 
    \caption{An example of the thought process for D2R in maze navigation task.}
    \label{fig:case2}
\end{figure*}

Figure~\ref{fig:case2} provides an example of the thought process of D2R in maze navigation task. In each step, D2R receives the original frame, then mark it with the current position. According to the marked frame and full video, D2R judges the dangerous position and generates a safe move direction for now, until it reaches the destination.

\end{document}